\documentclass[11pt]{article}
\pdfoutput=1
\usepackage{PRIMEarxiv}

\usepackage[utf8]{inputenc} 
\usepackage[T1]{fontenc}    
\usepackage{hyperref}       
\usepackage{url}            
\usepackage{booktabs}       
\usepackage{amsfonts}       
\usepackage{nicefrac}       
\usepackage{microtype}      
\usepackage{lipsum}
\usepackage{fancyhdr}       
\usepackage{graphicx}       
\graphicspath{{media/}}     
\usepackage{subcaption}
\usepackage{diagbox}
\usepackage{siunitx}
\usepackage{amsmath}
\usepackage{algpseudocode}
\usepackage{algorithm}
\usepackage{placeins}
\usepackage{amsfonts}
\usepackage{amssymb}
\usepackage{float}
\usepackage{multirow}
\usepackage{footnote}
\usepackage{pifont}

\DeclareMathOperator*{\argmax}{\arg\!\max}
\title{A Unified Image Preprocessing Framework For Image Compression}

\author{
  Moqi Zhang \\
  ByteDance Inc. \\
  Beijing, China\\
  \texttt{zhangmoqi@bytedance.com} \\
  \And
  Weihui Deng \\
  ByteDance Inc. \\
  Beijing, China \\
  \texttt{dengweihui@bytedance.com} \\
  \And
  Xiaocheng Li \\
  ByteDance Inc. \\
  Beijing, China \\
  \texttt{lixiaocheng.rd@bytedance.com} \\
}

\begin{document}
\maketitle

\begin{abstract}
With the development of streaming media technology, increasing communication relies on sound and visual information, which puts a massive burden on online media. Data compression becomes increasingly important to reduce the volume of data transmission and storage. To further improve the efficiency of image compression, researchers utilize various image processing methods to compensate for the limitations of conventional codecs and advanced learning-based compression methods. Instead of modifying the image compression oriented approaches, we propose a unified image compression preprocessing framework, called Kuchen, which aims to further improve the performance of existing codecs. The framework consists of a hybrid data labeling system along with a learning-based backbone to simulate personalized preprocessing. As far as we know, this is the first exploration of setting a unified preprocessing benchmark in image compression tasks. Results demonstrate that the modern codecs optimized by our unified preprocessing framework constantly improve the efficiency of the state-of-the-art compression.
\end{abstract}


\section{Introduction}
\label{sec:Introduction}
With the popularization of mobile devices and the development of information technology, information is increasingly interacted in the form of video, images and audios, etc. The transmission and storage of visual information bring great burden to streaming media. Therefore, reducing the irrelevance and redundancy of images and videos has become an active research topic in recent years. Many advanced image compression technologies play an important role in efficient information transmission. Existing algorithms can be classified into classical image compression standards such as WebP \cite{google}, JPEG\cite{pennebaker1992jpeg} and learning based image compression algorithms \cite{jiang2017end} \cite{toderici2017full} \cite{prakash2017semantic}. To the best of our knowledge, the main drawback of conventional compression standards is that they use a unified hand-craft encoder and decoder architecture, which limits their flexibility. Among them, a considerable amount of neural networks based approaches show its discernible talent compared to conventional image codecs\cite{jiang2017end} \cite{toderici2017full} \cite{prakash2017semantic}. 

As we know, during image capturing, compression, storage, and transmission, there will be non-negligible artifact distortion, among which noise, block effects, and blurring are more common. Therefore, choosing suitable image pre and postprocessing algorithms can alleviate this situation, improve the visual appearance, and even avoid information redundancy. To further improve image compression efficiency and obtain high-quality images, many methods of image preprocessing and postprocessing are widely used. Lots of approaches have been introduced in many papers, researchers not only use traditional computer vision image processing techniques\cite{yang2009image,chen2010image}, but also use deep learning based methods\cite{toderici2017full,prakash2017semantic}. However, different compression scenarios bring different compression artifacts. For instance, high-quality and high-resolution images need to remove information redundancy, and low-resolution blurred images need to enhance texture. Therefore, it is necessary to select the appropriate preprocessing method according to the characteristics of each image. Still, at the same time, it is difficult to find the optimal scheme that can balance the trade-off between image quality and bit rate. Blindly executing multiple image processing algorithms in series is time-consuming and may not achieve better results. Furthermore, there is no universal preprocessing and postprocessing method for image compression or a benchmark for processing selection. The main reason is that various factors have caused different problems, demonstrating that we need a robust and unified approach to solve these problems.

In this work, we proposed a hybrid preprocessing framework for image compression, which consists of a hybrid data labeling system and a learning-based image compression preprocessing model. In the data labeling framework, various image processing methods and one specific codec are used to create the annotated dataset as the training labels. In other words, we use neural networks to simulate different image processing methods concurrently and automatically choose an optimal solution for each pixel. The model trained on annotated data can be used as image preprocessing step and the outputs feed to different engineered codecs such as JPEG \cite{pennebaker1992jpeg} and H.265/HEVC \cite{sullivan2012overview} which is able to optimize the performance of the codecs. We present a comprehensive experiment to evaluate the performance of this framework on datasets of Douyin images and MIT300\cite{judd2012benchmark} respectively, including VQscore\cite{li2021full} and bits per pixel(BPP). By balancing trade-off between compression ratio and subjective quality preservation, the results demonstrate that our model consistently outperforms other image preprocessing algorithms.

In summary, our main contributions are as follows, 
\begin{itemize}
\item We propose a unified image compression processing framework to realize automatic image personalization, called Kuchen, which consists of a hybrid data labeling system and a convolutional neural network(CNN) backbone to capture the features from labels created by the former. 
\item We propose a hybrid data labeling system that selects the best modification by various image preprocessing methods along with one specific codec, which achieves the lowest bit rate without sacrificing subjective perception.
\item We formulate a compact and efficient CNN model to realize tailored image processing for each image. 
\end{itemize}

This paper is organized into 5 sections. Section \ref{sec:Introduction} provides a brief background and motivation of our work and also illustrates the concise structure of the paper. Section \ref{sec:related_work} demonstrates several typical modern algorithms and state-of-the art learning based models in the field of image processing and image compression. The unified image processing framework is presented and evaluated in Section \ref{sec:Methodology} and \ref{sec:Experiment}, respectively. We also comprehensively compared the performance of traditional engineered codecs with or without our preprocessing model in the last part of Section \ref{sec:Experiment}. Finally, we summarized the key contributions of our study, then detailed the limitations of our framework and discuss the exploration direction of future work in Section \ref{sec:conclusion}.

\section{Related Work}
\label{sec:related_work}
\subsection{Learning Based Image Compression}
\label{sec:Compression}
Image compression can be classified into lossy and lossless compression, while the former is a mainstream approach to reduce the volume of image transmission and storage. Recent papers have revealed the great potential of deep neural networks and taken a further step towards better visual quality compression, which received increasing attention \cite{jiang2017end,toderici2017full,prakash2017semantic,toderici2015variable,mentzer2018conditional}. In particular, some end-to-end image compression architecture outperformed modern codecs and achieved competitive performance. Since Google first applied LSTM to image compression, its PSNR performance has been superior to JPEG, JPEG2000, and WebP\cite{toderici2017full} \cite{toderici2015variable}. Then motivated by the success of SRCNN in image resolution tasks, Feng Jiang et al. extended convolutional neural networks(CNN) to compression schemes. He proposed ComCNN and RecCNN, which correspond to the encoding and decoding, respectively, and achieved impressive performance \cite{jiang2017end}.

Meanwhile, Fabian et al. first trained a lossy image compression auto-encoder to work with a context model \cite{mentzer2018conditional}. The auto-encoder uses the context model to estimate entropy and controls the representation's trade-off between reconstruction distortion and entropy. Aaditya et al.pointed out the power of semantic perception when doing image compression by adopting a human perspective to understand images, then encoding the region of interest(ROI) with a higher bit rate and lowering the bit rate elsewhere in the image \cite{prakash2017semantic}.

Generative lossy compression systems have become popular due to the high-quality reconstruction that perfectly balances the "rate-distortion-perception" trade-off \cite{mentzer2020high,agustsson2019generative}. In \cite{agustsson2019generative}, E. Agustsson et al. formulate a full-resolution image compression framework to obtain extreme compressed images with a bit rate below 0.1 BPP. In \cite{judd2012benchmark}, the generative model High Fidelity Compression (HiFiC) is able to obtain high perceptual fidelity output very similar to the input, including detailed textures such as human skin.

\subsection{Image Processing}
\label{sec:Processing}
Image processing refers to image analysis and processing, consisting of high, mid, and low levels. In the field of image compression, generally, we focus on low-level vision problems, including denoising, and image enhancement. To optimize the performance of image compression,  commonly used processing algorithms can be classified into image restoration and quality enhancement. In the image restoration domain, the various filters proposed to suppress impulse noise \cite{chen1999tri,chen2006new,buades2011non}. In \cite{chen1999tri}, they propose a novel nonlinear median filter and apply it to the image conditionally depending on the results of the standard median (SM) filter and the center-weighted median (CWM) filter. Meanwhile, the paper \cite{chen2006new} proposed a weighted median filter according to the local variance of the image. In the paper \cite{buades2011non}, an improved weighted median filtering algorithm is proposed to obtain the denoised image by classifying and weighted averaging the regions with similar pixels. Over the past several years, there have been numerous learning-based studies to cope with noise and blurring problems and go beyond traditional filter-based approaches into the mainstream. More famously, DnCNNs is able to handle Gaussian denoising \cite{zhang2017beyond}, and SSDA is designed to combine sparse coding, and deep networks \cite{xie2012image}. To deal with image restoration tasks, convolutional networks show their talents in complex tasks due to the same size of network input and output, and related papers are increasingly sprung up \cite{liu2018multi,jia2021ddunet}.

Despite the color of images, histogram equalization is widely used for image enhancement. Moreover, Original histogram equalization, automatic white balance, and automatic color equalization are also commonly used by processing the Red, Green, and Blue channels separately \cite{hummel1975histogram}. White balance often disregards the fixed percentage of pixels at the end of histograms. Unlike histogram equalization, color equalization converts RGB format into HSI format(Hue, Saturation, and Intensity) and enhances the intensity channel. In the paper \cite{gangkofner2007optimizing}, a classical high-pass filter is proposed to enhance the details and edge information of the image. On the other hand, image enhancement tends to improve the quality of the image in poor conditions and obtain a better representation. For human viewers, artifact removal techniques work with regard to image enhancement. Blocking, ringing effects, and blurring are the most common image artifacts, dramatically reducing human experiences. The state-of-the-art approaches are neural network-oriented methods \cite{svoboda2016compression,yu2016deep,galteri2017deep}. Like \cite{dong2014learning}, the paper \cite{svoboda2016compression} propose a compression artifact reduction approach using a compact convolutional neural network. To cope with various compression artifacts, \cite{yu2016deep} formulate a new deep CNN and significantly improve acceleration strategies. In \cite{galteri2017deep}, Leonardo Galteri removes the artifacts using a conditional generative adversarial network. 

We present one more algorithm as an image processing method to reduce the bit rate in this study, which is Google's codec Guetzli. Guetzli is an improved JPEG encoder that uses butteraugli as a benchmark to obtain human-perceivably indistinguishable images at a lower bit rate than the original JPEG encoder \cite{alakuijala2017guetzli}. One of our bold assumptions is that there is great potential for applying different modern codecs simultaneously, eliminating various image artifacts and information redundancy. Overall, image processing algorithms have been comprehensively investigated for various purposes. We explore how to combine them and personalize the images according to their characteristics.

\section{Methodology}
\label{sec:Methodology}
In this section, we introduce the whole framework architecture we explore for end-to-end image compression. Inspired by Google's Model Soups \cite{wortsman2022model} and recent learning-based advances in image compression \cite{jiang2017end,toderici2017full,prakash2017semantic,judd2012benchmark,toderici2015variable,mentzer2018conditional,mentzer2020high}, we utilize neural networks to capture the characteristics of various image processing concurrently and adaptively realize images processing, which can be regarded as Image Processing Soup. Before having a comprehensive understanding of our implementation, we can take a look at the implementation of Model Soup. Google's research group finetune the pre-trained model with different hyperparameters, then average weights of multiple models to substitute into the final model. Encouraged by its impressive success, we extend this idea to compression preprocessing schemes, perform various image processing oriented methods to train data respectively, then pick the individual processed image with the highest performance as the training label. Undoubtedly, they provided a concise and robust model soups approach that achieved a new state-of-the-art. Through analogy learning, we extend this fine-tuning of the average idea to image processing and propose a unified preprocessing framework for image compression as follows.

\subsection{Architecture of End-to-End compression Framework}
\begin{figure}[ht]
  \centering
  \includegraphics[width=0.9\textwidth]{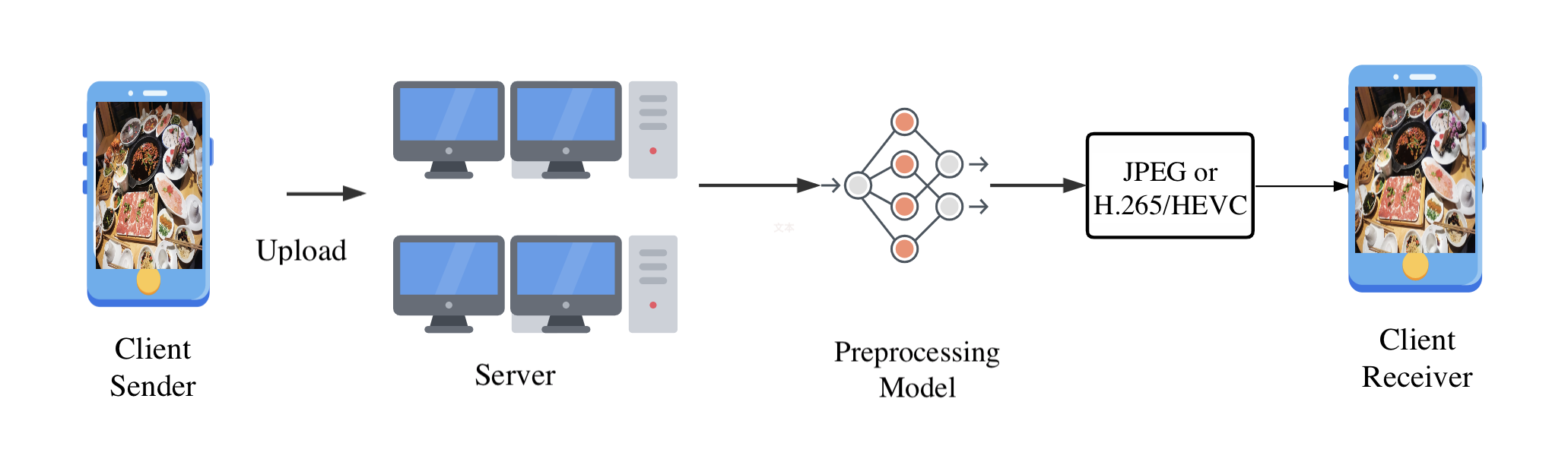}
  \caption{End-to-End compression Framework}
  \label{fig:framework}
\end{figure}

We present an efficient image compression preprocessing network that incorporates multiple image preprocessing algorithms and adaptively implements personalized processing to a particular image—the sample image compression framework as shown in fig. \ref{fig:framework}. Client uploads the raw images to the server as source materials, but these images will be distorted since the processing of the capturing environment and the built-in codecs of the devices. To eliminate noise and unwanted information in images during vision communication, the live streaming servers utilize different image codecs to realize different levels of image compression in various scenes. Before sending back images to the receiver, our preprocessing model plays a significant role in this regard, reducing image compression distortion, blur or noise effects to enhance image texture information, and aiming to reduce spatial redundancy and improve the compression performance of codecs while maintaining the user experience. It applies tailored image processing to the current image and feeds the preprocessed image into the original codec for compression work. The image received by the end-users will be at a low bit rate without decreasing the subjective perception, some of which are well suited to human viewers.

\FloatBarrier
\label{sec:compression Framework}
\begin{figure}[ht]
  \centering
  \includegraphics[width=\textwidth]{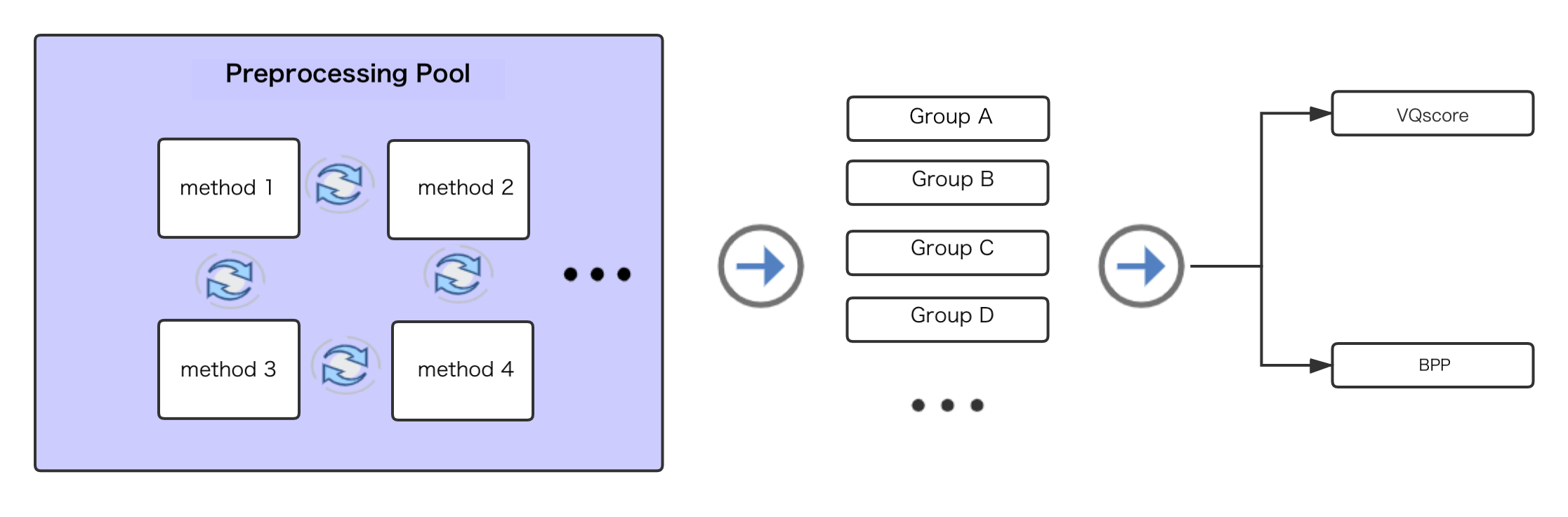}
  \caption{Data Labeling Framework}
  \label{fig:labeling}
\end{figure}
\FloatBarrier
\subsection{Architecture of Data Labeling Framework}
\label{sec:Labeling Framework}
We propose preprocessing techniques to maximize the compression efficiency of codecs and provide a better representation of images for visual communication. A novel image preprocessing labeling framework is the cornerstone of this work and can be seen as fig. \ref{fig:labeling}. Our method allows us to fit multiple image processing algorithms concurrently and can also effectively accommodate standard codecs. The innovation point of this method is the data labeling framework. According to different image preprocessing purposes, a variety of advanced image processing methods are selected as candidates in the preprocessing pool, including method 1 and method 2 and so on. In data annotation, we use the equation of permutation \ref{eqn:permutations} to apply the combined groups to raw images, where the combined groups can be divided into A, B, C, and other groups. For example, group A is method 1 and method 2 in series, and group B is method 1 and method 3 in series.

Equation of $k$-permutations of $n$ \cite{enwiki:1102976887},
\begin{equation}
\label{eqn:permutations}
P(n, k) = (n) * (n - 1) * (n - 2) * \cdots * (n - k + 1) \\
= \frac{n!}{n - k!}
\end{equation}
where $n$ is the number of processing methods and $k$ is the number of processing method in a combination group.

To sum up, we consider multiple image processing methods and ignore the effects of the processing sequence, which constitute a total of bunch groups of processed labels(A, B, C, AB, AC, etc.). We modify every element in the training set through these groups of combinations respectively, then calculate the score of each of them according to equation \ref{eqn:score}. 

\begin{equation}
\label{eqn:score}
Score(img_{i,j}) = Z(VQscore(img_{i,j})) + Z(BPP(img_{i,j})
\end{equation}

\begin{equation}
\label{eqn:VQscore}
Z(VQscore(img_{i,j})) = \frac{VQscore(img_{i,j}) - mean(VQscore(img_i))}{\sigma}
\end{equation}

\begin{equation}
\label{eqn:BPP}
Z(BPP(img_{i,j})) = \frac{BPP(img_{i,j} - mean(BPP(img_i))}{\sigma}
\end{equation}

\begin{equation}
\label{eqn:mean_VQscore}
mean(VQscore(img_i)) = \frac1{n}\sum_{\substack{0<j<n}}VQscore(img_{i,j})
\end{equation}

\begin{equation}
\label{eqn:mean_BPP}
mean(BPP(img_i))
= \frac1{n}\sum_{\substack{0<j<n}}BPP(img_{i,j})
\end{equation}

Where $i$ denotes the particular number of images, $n$ denotes the total number of the preprocessing groups, $j$ denotes the particular number of preprocessing groups, and $Z$ function standardizes the VQscore and BPP to balance the trade-off between visual quality and compression ratio. According to equations \ref{eqn:VQscore}, \ref{eqn:BPP}, \ref{eqn:mean_VQscore}, \ref{eqn:mean_BPP}, it scales the VQscore and BPP, which sets their mean and variance to zero and 1, respectively. In order to eliminate the influence of different dimensions of feature, it is an essential step to do data standardization in feature processing.

The score of each image is derived from the sum of VQscore and BPP, which takes both image quality and storage space into equal consideration. We build labels corresponding to the training data by selecting the image with the highest score. The entire workflow of our data labeling system algorithm can be summarized as algorithm \ref{alg:cap}.
\begin{algorithm}
\caption{Data Labeling Algorithm}
\label{alg:cap}
\begin{algorithmic}
\Require{$img_{1} \dots img_{N}$}
\Ensure{$label_{1} \dots label_{N}$}
\State {$N$ $\gets$ {$length(Imgs)$}}
\State {$M$ $\gets$ {$length(Groups)$}}\footnotemark[1]{}
\For{\texttt{$i \gets 1$ to $N$}}
    \For{\texttt{$j \gets 1$ to $M$}}
        \State {$img_{i,j}$ $\gets$} \texttt{Apply <$Group_j$, Codecs> to $img_i$} \footnotemark[1]{}
        \State \texttt{$Score(img_{i,j}) = Z(VQscore(img_{i,j})) + Z(BPP(img_{i,j})$}
      \EndFor
      \State $<i, j^{\star}> = \argmax{Score(img_{i,j})}$
      \State $label_{i} \gets img_{i, j^{\star}}$ 
\EndFor
\end{algorithmic}
\end{algorithm}
\footnotetext[1]{$N$ denotes the total number of the images in training set, $M$ denotes the total number of the combined preprocessing groups. <$Group_j$, Codecs> denotes applying the preprocessing group $Group_j$ to the particular image and then feeding the output to a specific codec, such as JPEG.}

\subsection{Architecture of Neural Network}
\label{sec:Neural Network}

\begin{figure}[ht]
  \centering
  \includegraphics[width=\textwidth]{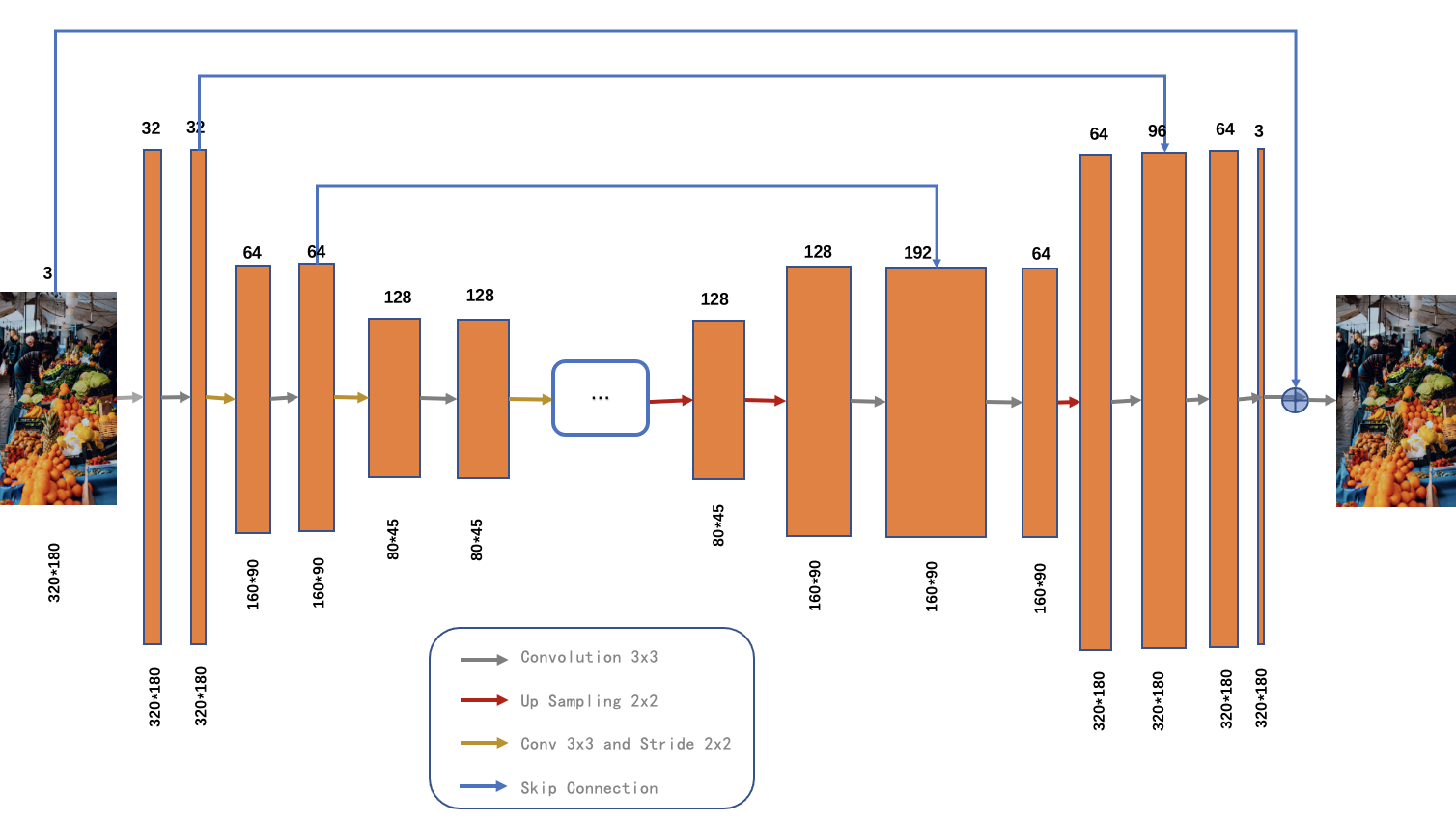}
  \caption{Architecture of Neural Network}
  \label{fig:network_architecture}
\end{figure}

The preprocess network architecture is also a principal element in our framework. Our proposed network, Kuchen, is based on the popular vision tasks model—U-net\cite{ronneberger2015u}, and its image denoising oriented version—DDUNet\cite{jia2021ddunet}. As shown in fig. \ref{fig:network_architecture}, Kuchen consists of a three-level (three downsampling layers and corresponding three-level upsampling) U-net layers with one global residual structure. Each downsampling level is composed of one $3\times3$ convolution with the stride of 2(except the first level) followed by a typical $3\times3$ convolutional layer with a double feature channel. Like the downsampling level, each upsampling level consists of one $2\times2$ up-convolution layer followed by one halve-channel $3\times3$ convolutional layer and one typical $3\times3$ convolution.

Consequently, the input and output of the framework are the same sizes. The last layer of each downsampling level concatenates with the same size halve-channel convolution layers, further enhancing each level's key features. In terms of global residual structure, it connects the input with the output of the last layer in the U-net by adding them together, which pretends the unwanted information as the residual to reduce the convergence time of the model. According to the equation \ref{eqn:I}, the output of the whole network aims to reconstruct a clear and low bit rate image $i_{new}$ by removing the noise and information redundancy $i_{diff}$ from original image $i_{ori}$.

\begin{equation}
\label{eqn:I}
I_{new} = I_{ori} - I_{diff}
\end{equation}

\begin{equation}
\label{eqn:MSE}
MSE = \frac{1}{M*N}\sum_{\substack{0<m<M}}\sum_{\substack{0<n<N}}[I_{new} - I_{label}]^2
\end{equation}

With the help of the residual structure, not only the convergence time and the number of parameters has been effectively reduced, but also the behavior of encoding and decoding has been improved. In this study, we use the standard mean squared error (MSE) objective function as the loss function during the training step as equation \ref{eqn:MSE}, which is generally used in image reconstruction tasks to measure the changes between input and output images. 

\section{Experiment}
\label{sec:Experiment}
In this section, we demonstrate comprehensive experiments to evaluate the performance of with or without our preprocessing model in image compression. All the experiment results were computed on Bytedance Inc.'s Douyin Dataset and MIT300 dataset \cite{judd2012benchmark,mit-tuebingen-saliency-benchmark}. 

\subsection{Perceptual Metrics}
\label{sec:Perceptual Metrics}
The conventional recipe of perceptual metrics are MS-SSIM and PSNR, which are used to evaluate the quality of compressed images \cite{jiang2017end,toderici2017full,prakash2017semantic,toderici2015variable,mentzer2018conditional,mentzer2020high} and estimate the image distortion rate. It is a common issue in streaming media; these full-reference perceptual metrics assume that the original video was not pristine and manipulated (processed, compressed, etc.). Nevertheless, evaluating the quality of preprocessed images is a major issue for most of the existing full-reference quality metrics. Filtered images remove information redundancy and noise in the original image. Consequently, these changes make the image look less similar to the original image, leading to the deterioration of full-reference evaluation metrics such as PSNR, but look better to the human viewers. In this case,  we assess our framework on no-referenced image quality assessment, VQscore, consisting of a CNN backbone for feature extraction and a transformer encoder for quality prediction \cite{9746539,li2021full}. VQscore is well-tailored to enhanced and filtered UGC and outperforms the other encoding-quality assessment systems in Challenge on Quality Assessment of Compressed UGC Videos ICME 2021 \cite{ICME}, which is more suitable for actual application scenarios, and more similar to users with aesthetic preferences. VQscore can be used to evaluate the quality of all references (FR) and no reference (NR), but we only consider the NR score since the end-users of streaming media only care about whether the visual experience of the current image is satisfactory.

\subsection{Dataset}
\label{sec:Dataset}
All experiments are tested on two datasets, Bytedance Inc.'s Douyin Dataset and MIT300 dataset, respectively. The Douyin dataset contains 2183 source images from Douyin App with a variety of image content. The MIT dataset consists of 300 source images with natural indoor and outdoor scenes. The images are not pristine, and some have been processed with special effects, such as beautification. In this work, we split the data between the training and testing sets in around 0.6:0.4 ratio(Douyin) and 0.8:0.2 ratio(MIT). Due to the limitation of the number of images in MIT300 dataset, we keep 240 images for training steps. During training, before feeding into models, all input images are sliced into fixed-size patches (320*180*3) with a stride of 160 pixels as data augmentation, limiting the number of model parameters and augmenting the amount of training data. Therefore, the training set of the Douyin dataset is about 30k, and the MIT300 is around 6k.

In terms of modern codec, we considered three compression standards to compress the
pre-processed images: JPEG \cite{pennebaker1992jpeg}, H.265/HEVC \cite{sullivan2012overview} and WebP \cite{google}. 

In total four image processing methods are considered for this study, including the traditional image codec Guetzli \cite{alakuijala2017guetzli}, a compression distortion removal model (AR) proposed by the Media Foundation Team, Bytedance Inc., traditional image denoising method implemented by OpenCV(Fast Denoising), and texture enhancement method implemented by Python Imaging Library(Detail Filter). Fast Denoising is proposed in \cite{buades2011non} called non-local means denoising, which selects a template patch and replaces the pixels in it by averaging all similar neighborhood windows. Meanwhile, the Detail Filter method is a classical approach to applying high pass filters to images.

\subsection{Experiment Result}
\label{sec:Experiment Result}
\begin{figure}[ht]
  \centering
  \includegraphics[width=\textwidth]{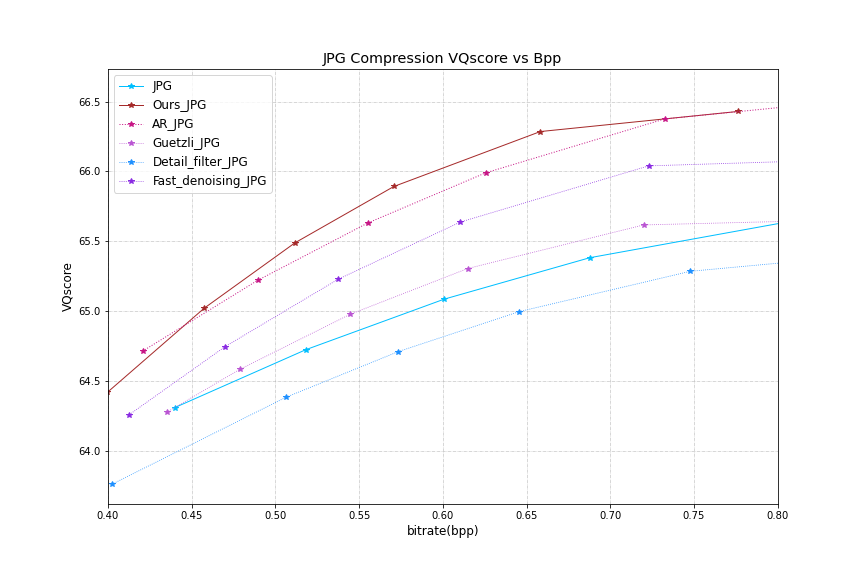}
  \caption{VQscore vs. BPP}
  \label{fig:VQscore_BPP}
\end{figure}
In this section, we set different quality factors (QF) to make all of the experimental group approach the same bit per pixel. We trained our preprocessing model using an Adam training optimizer \cite{kingma2014adam} with an initial learning rate of $1\times10^{-4}$ and batch size of 4.
As shown in fig. \ref{fig:VQscore_BPP}, we evaluated the compression performance of JPEG, including visual quality(VQscore) and BPP, with or without our preprocessing model compared to the commonly used image processing algorithms. The image compression framework with our model maintains and even improves the image quality, especially when the BPP is between a range of 0.50-0.70. Not surprisingly, when the BPP is in the medium range, our model significantly outperforms other algorithms, and in the rest of the cases(BPP lower than 0.45 and BPP higher than 0.75), the superiority is not obvious, even similar to the controlled experiments. It could be construed as an extreme compression scenario; codecs weaken the subtle changes that preprocessing makes to images. In particular, although one of our participants, a simple denoising algorithm, performed worse than the JPEG baseline, it is still worth considering because it may works in some corner cases. 

\begin{figure}[ht]
  \centering
  \includegraphics[width=\textwidth]{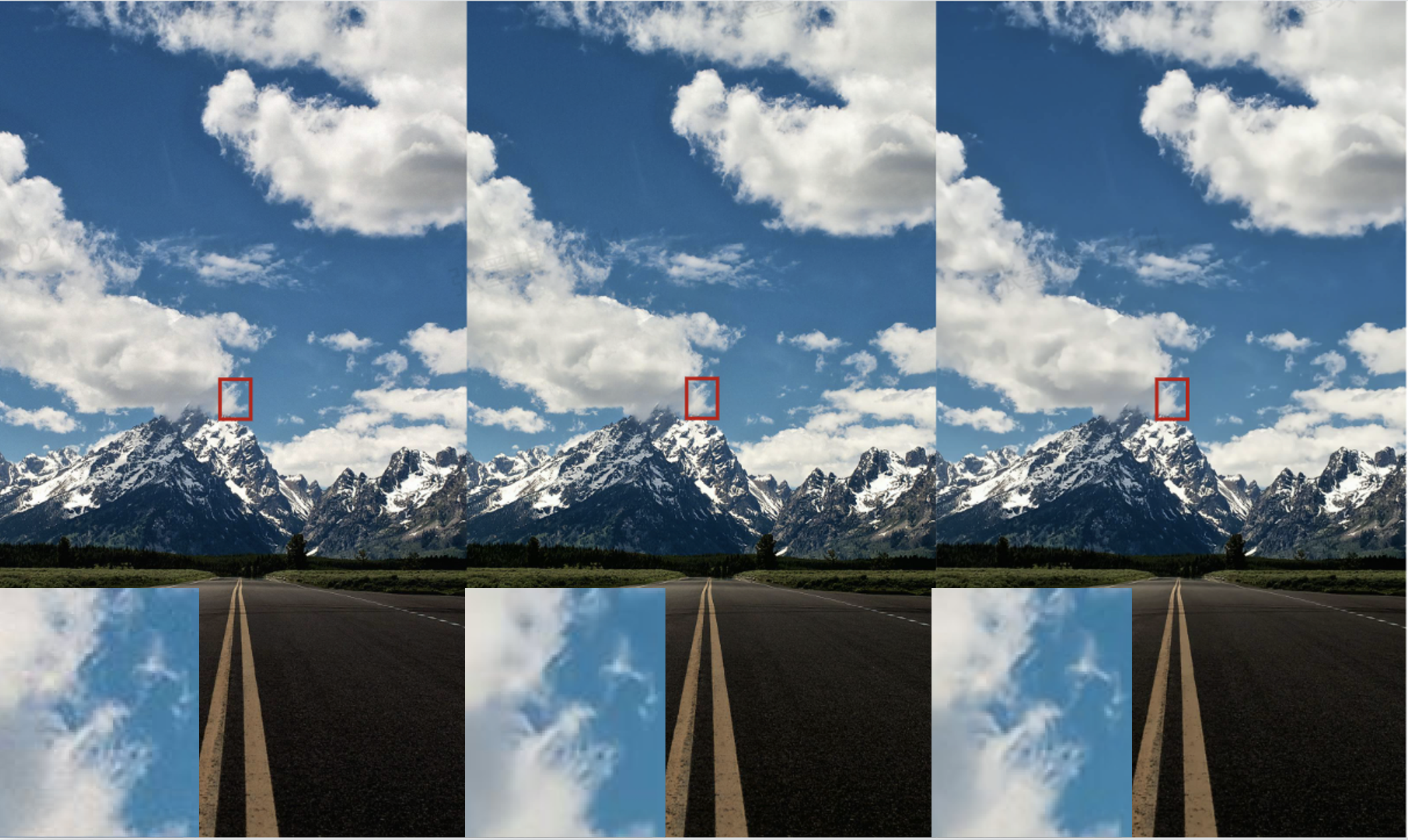}
  \caption{Original(left) vs. AR(medium) vs. Ours(right). Comparison of compression of landscape image in standard JPEG with different preprocessing methods. File sizes of them are 277KB, 298KB, and 216KB(ours) respectively.}
  \label{fig:compared}
\end{figure}

\begin{figure}[ht]
\begin{subfigure}{.33\textwidth}
  \centering
  \includegraphics[width=0.9\linewidth]{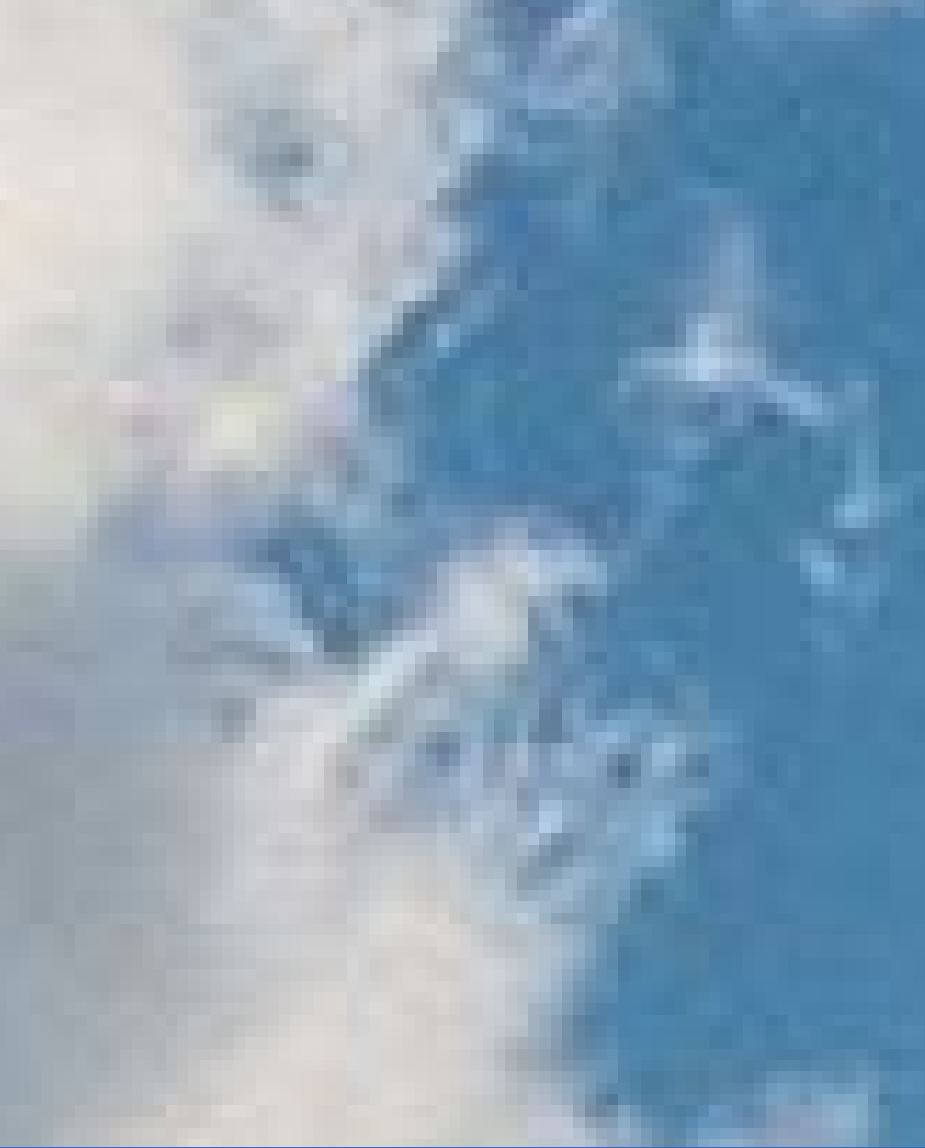}
  \caption{Original}
  \label{fig:sfig1}
\end{subfigure}%
\begin{subfigure}{.33\textwidth}
  \centering
  \includegraphics[width=0.9\linewidth]{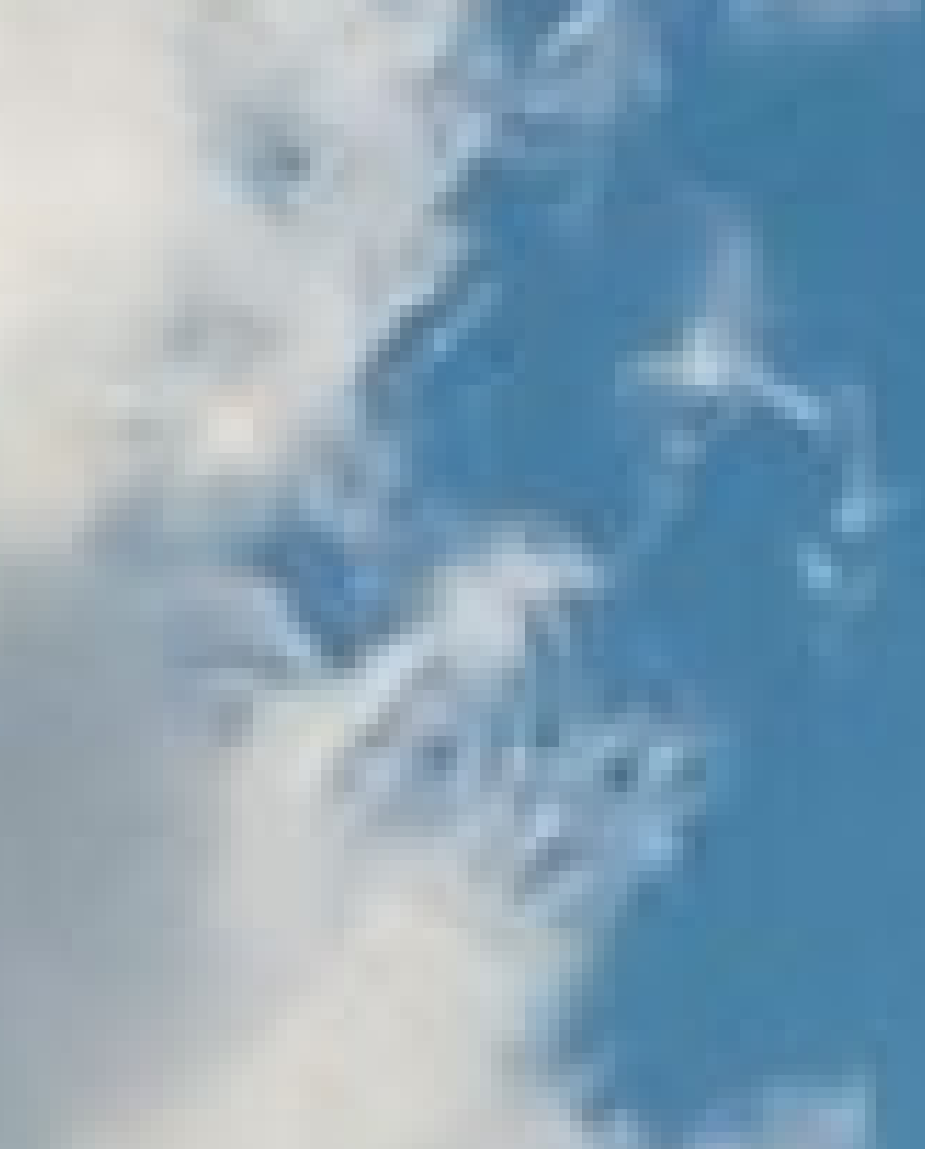}
  \caption{AR}
  \label{fig:sfig2}
\end{subfigure}
\begin{subfigure}{.33\textwidth}
  \centering
  \includegraphics[width=0.9\linewidth]{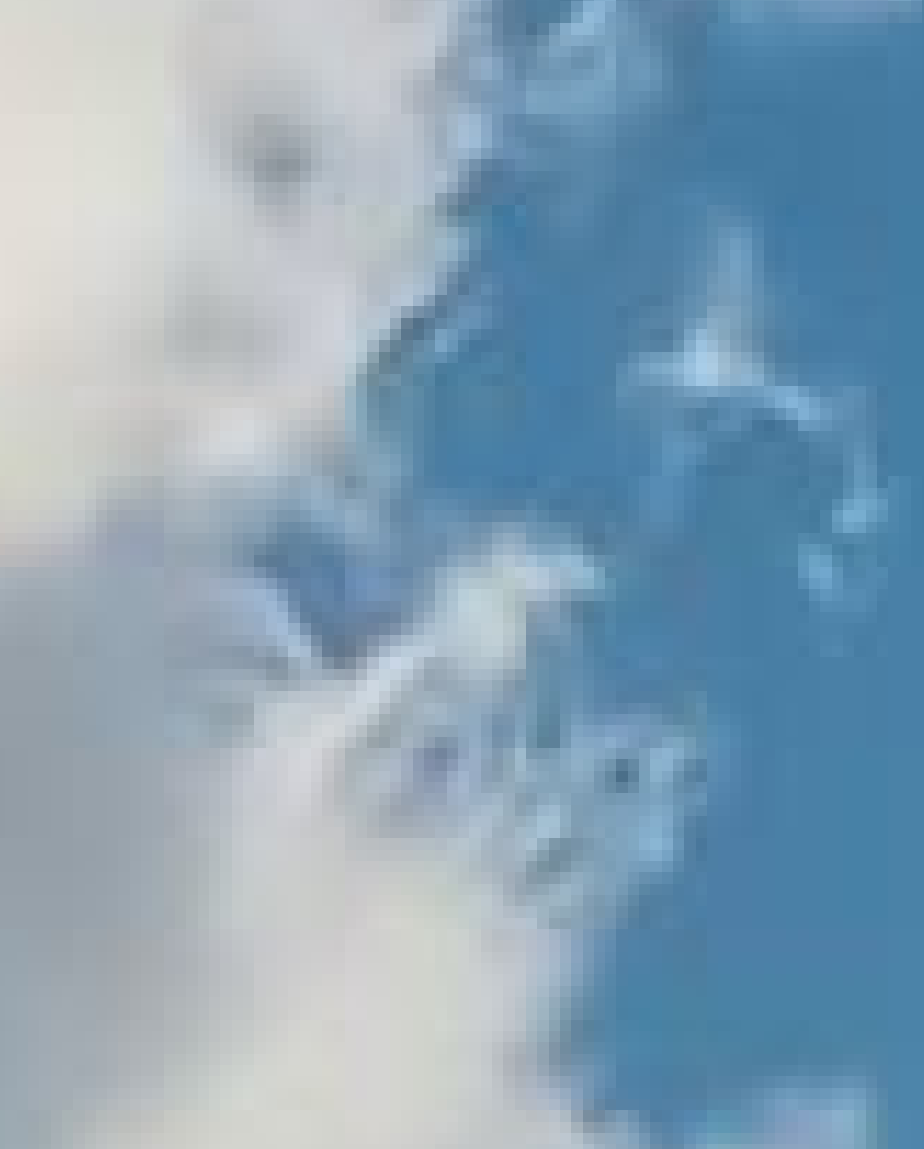}
  \caption{Ours}
  \label{fig:sfig3}
\end{subfigure}%
\caption{Original(a) vs. AR(b) vs. Ours(c). Detailed texture comparison of compression of landscape image.}
\label{fig:zoomin}
\end{figure}

According to fig. \ref{fig:VQscore_BPP}, the largest difference of VQsocre among the various processing algorithms is smaller than 1.5. In order to show how these differences affect images visual perception, we visualise a sample JPEG compressed landscape photo using three kinds of approaches shown in fig. \ref{fig:compared}. It is hard to discern the difference with the naked eye unless you zoom in to a certain size to identify differences in pixels. Fig. \ref{fig:zoomin} shows the detailed texture results on the landscape image from Douyin Dataset. The reconstruction image by JPEG is left, where we can see the blocking artifacts at the edges of clouds. The middle and right reconstruction images have been preprocessed by AR and our model, respectively, where we can see the edges of clouds are smoother and clear. All in all, our preprocessing framework provides a better representation of images at lower bit rates.

The afterward results show the potential of a hybrid image preprocessing framework in image compression tasks, but the architecture of the training model is replaceable. In order to demonstrate that our hybrid image processing model can be used for different compression codecs, we also tabulated the performance of three codecs with different pre-processing methods on the Douyin Dataset in table \ref{tab:douyin}. Among these three group experiments, our model is a crucial demonstration of value for compression with better visual quality and lower bit rate. As shown in table \ref{tab:douyin}, the success of our model is not limited to the choice of codecs. Among these three codecs, H.265/HEVC obtains quite a better result, achieving 34\% compression ratio improvement without dramatically affecting the visual quality of images. Meanwhile, in terms of JPEG and WebP, our model still achieves 22.65\% and 24.75\% improvement in compression ratio. Comparing the performance of our model with the rest in the processing pool, the VQscore of our model is almost the same as the result of AR but much higher than the other three. Nevertheless, at the same time, our model dramatically improves the compression ratio via standard codecs, which is more than 10\% higher than other methods.

\FloatBarrier
\FloatBarrier

\FloatBarrier
\begin{table}[hbt!]
\begin{center}
\caption{Douyin Dataset}
\renewcommand{\arraystretch}{1.4}%
\begin{tabular}{c|c|c|c|c|c|c|c}
\hline
\hline
\multicolumn{2}{|c|}{\diagbox{Codecs}{Preprocessing}} & Original & Kuchen  & AR  & Guetzli  & Detail Filter & Fast Denoising \\ \hline
\multirow{3}{*}{JPEG} & File Size (MB) & 267.21 & 206.67 & 234.24 & 235.70 & 234.39 & 231.14 \\ \cline{2-8} 
 & Compression Ratio & N/A & \textbf{22.65\%} & 12.34\% & 11.79\% & 12.28\% & 13.50\% \\ \cline{2-8} 
 & VQscore & 65.75 & \textbf{66.48} & \textbf{66.62} & 65.71 & 65.47 & 66.16 \\ \hline
\multirow{3}{*}{HEVC} & File Size (MB)& 118.63 & 77.76 & 100.61 & 107.20 & 87.87 & 104.09 \\ \cline{2-8} 
 & Compression Ratio & N/A & \textbf{34.45\%} & 15.19\% & 9.63\% & 25.93\% & 12.25\% \\ \cline{2-8} 
 & VQscore & 65.64 & \textbf{66.11} & \textbf{66.18} & 65.55 & 65.30 & 65.87 \\ \hline
\multirow{3}{*}{WebP} & File Size (MB)& 136.08 & 102.39 & 119.42 & 125.39 & 105.64 & 117.39 \\ \cline{2-8} 
 & Compression Ratio & N/A & \textbf{24.75\%} & 12.24\% & 7.85\% & 22.36\% & 13.73\% \\ \cline{2-8} 
 & VQscore & 65.34 & \textbf{65.71} & \textbf{65.86} & 65.24 & 64.95 & 65.50 \\ \cline{1-8} 
 \hline
 \hline
\end{tabular}
\label{tab:douyin}
\end{center}
\end{table}
\FloatBarrier
\FloatBarrier
\begin{table}[hbt!]
\begin{center}
\caption{MIT300 Dataset}
\renewcommand{\arraystretch}{1.5}%
\begin{tabular}{c|c|c|c|c|c|c|c}
\hline
\hline
\multicolumn{2}{|c|}{\diagbox{Codecs}{Preprocessing}} & Original & Kuchen & AR  & Guetzli  & Detail Filter & Fast Denoising \\ \hline
\multirow{3}{*}{HEVC} & File Size (MB) & 4.68 & 3.57 & 3.92 & 4.18 & 3.77 & 4.08 \\ \cline{2-8} 
 & Compression Ratio & N/A & \textbf{23.76\%} & 16.19\% & 10.70\% & 19.44\% & 12.92\% \\ \cline{2-8} 
 & VQscore & 63.80 & \textbf{63.95} & \textbf{64.16} & 62.67 & 60.94 & \textbf{63.96} \\ \hline
 \hline
\end{tabular}
\label{tab:MIT}
\end{center}
\end{table}
\FloatBarrier

Most images in Douyin dataset are high-resolution images about 1080p and above. To further demonstrate the effectiveness of our model in a full-resolution scenario, we evaluate the performance of the classical public dataset, MIT300 in table \ref{tab:MIT}. The results report that Kuchen always provides the smallest file size with acceptable subjective quality. Unlike the results in table \ref{tab:douyin}, the compression ratio improvement by our model is 23.76\%, which is not very huge since the resolution is smaller than 720p. This indicates that our framework is more suitable for high-resolution images. In future work, to improve the overall performance of full-resolution scenarios, we can explore other applicable advanced preprocessing algorithms as candidates.

\section{Conclusion}
\label{sec:conclusion}
The research of image processing has reached a certain extent, but the final performance by blindly using a variety of image processing approaches is not necessarily good. In this paper, we propose a unified image preprocessing framework to improve the compression ratio of the codec without dramatically sacrificing the visual quality of images. Our framework takes into account both the perspective experience of end-users and the efficiency of visual information transmission. The framework consists of data labeling algorithm and hybrid preprocessing model. According to the experiments, our model is notable for outperforming the other preprocessing algorithms in some scenarios. Furthermore, our goal is to propose a new exploration direction in terms of image compression, as well as in terms of image processing as a whole. Our unified workflow solves the complex problems of integrating various preprocessing methods, including long processing time, low efficiency, and performance conflicts.

In future work, we aim to extend our strategy into other vision schemes and in terms of language tasks. We believe that by carefully selecting the suitable preprocessing approaches and improving the complexity of the model, research in the field of image compression will be further advanced, and hybrid processing techniques are trending.



\begin{thebibliography}{10}

\bibitem{google}
An image format for the web.
\newblock Available at \url{https://developers.google.com/speed/webp/}.

\bibitem{pennebaker1992jpeg}
William~B Pennebaker and Joan~L Mitchell.
\newblock {\em JPEG: Still image data compression standard}.
\newblock Springer Science \& Business Media, 1992.

\bibitem{jiang2017end}
Feng Jiang, Wen Tao, Shaohui Liu, Jie Ren, Xun Guo, and Debin Zhao.
\newblock An end-to-end compression framework based on convolutional neural
  networks.
\newblock {\em IEEE Transactions on Circuits and Systems for Video Technology},
  28(10):3007--3018, 2017.

\bibitem{toderici2017full}
George Toderici, Damien Vincent, Nick Johnston, Sung Jin~Hwang, David Minnen,
  Joel Shor, and Michele Covell.
\newblock Full resolution image compression with recurrent neural networks.
\newblock In {\em Proceedings of the IEEE conference on Computer Vision and
  Pattern Recognition}, pages 5306--5314, 2017.

\bibitem{prakash2017semantic}
Aaditya Prakash, Nick Moran, Solomon Garber, Antonella DiLillo, and James
  Storer.
\newblock Semantic perceptual image compression using deep convolution
  networks.
\newblock In {\em 2017 Data Compression Conference (DCC)}, pages 250--259.
  IEEE, 2017.

\bibitem{yang2009image}
Ching-Chung Yang.
\newblock Image enhancement by the modified high-pass filtering approach.
\newblock {\em Optik}, 120(17):886--889, 2009.

\bibitem{chen2010image}
Qian Chen and Dapeng Wu.
\newblock Image denoising by bounded block matching and 3d filtering.
\newblock {\em Signal Processing}, 90(9):2778--2783, 2010.

\bibitem{sullivan2012overview}
Gary~J Sullivan, Jens-Rainer Ohm, Woo-Jin Han, and Thomas Wiegand.
\newblock Overview of the high efficiency video coding (hevc) standard.
\newblock {\em IEEE Transactions on circuits and systems for video technology},
  22(12):1649--1668, 2012.

\bibitem{judd2012benchmark}
Tilke Judd, Fr{\'e}do Durand, and Antonio Torralba.
\newblock A benchmark of computational models of saliency to predict human
  fixations.
\newblock {\em MIT-CSAIL-TR-2012-001}, 2012.

\bibitem{li2021full}
Yang Li, Longtao Feng, Jingwen Xu, Tao Zhang, Yiting Liao, and Junlin Li.
\newblock Full-reference and no-reference quality assessment for compressed
  user-generated content videos.
\newblock In {\em 2021 IEEE International Conference on Multimedia \& Expo
  Workshops (ICMEW)}, pages 1--6. IEEE, 2021.

\bibitem{toderici2015variable}
George Toderici, Sean~M O'Malley, Sung~Jin Hwang, Damien Vincent, David Minnen,
  Shumeet Baluja, Michele Covell, and Rahul Sukthankar.
\newblock Variable rate image compression with recurrent neural networks.
\newblock {\em arXiv preprint arXiv:1511.06085}, 2015.

\bibitem{mentzer2018conditional}
Fabian Mentzer, Eirikur Agustsson, Michael Tschannen, Radu Timofte, and Luc
  Van~Gool.
\newblock Conditional probability models for deep image compression.
\newblock In {\em Proceedings of the IEEE Conference on Computer Vision and
  Pattern Recognition}, pages 4394--4402, 2018.

\bibitem{mentzer2020high}
Fabian Mentzer, George~D Toderici, Michael Tschannen, and Eirikur Agustsson.
\newblock High-fidelity generative image compression.
\newblock {\em Advances in Neural Information Processing Systems},
  33:11913--11924, 2020.

\bibitem{agustsson2019generative}
Eirikur Agustsson, Michael Tschannen, Fabian Mentzer, Radu Timofte, and Luc~Van
  Gool.
\newblock Generative adversarial networks for extreme learned image
  compression.
\newblock In {\em Proceedings of the IEEE/CVF International Conference on
  Computer Vision}, pages 221--231, 2019.

\bibitem{chen1999tri}
Tao Chen, Kai-Kuang Ma, and Li-Hui Chen.
\newblock Tri-state median filter for image denoising.
\newblock {\em IEEE Transactions on Image processing}, 8(12):1834--1838, 1999.

\bibitem{chen2006new}
Jingdong Chen, Jacob Benesty, Yiteng Huang, and Simon Doclo.
\newblock New insights into the noise reduction wiener filter.
\newblock {\em IEEE Transactions on audio, speech, and language processing},
  14(4):1218--1234, 2006.

\bibitem{buades2011non}
Antoni Buades, Bartomeu Coll, and Jean-Michel Morel.
\newblock Non-local means denoising.
\newblock {\em Image Processing On Line}, 1:208--212, 2011.

\bibitem{zhang2017beyond}
Kai Zhang, Wangmeng Zuo, Yunjin Chen, Deyu Meng, and Lei Zhang.
\newblock Beyond a gaussian denoiser: Residual learning of deep cnn for image
  denoising.
\newblock {\em IEEE transactions on image processing}, 26(7):3142--3155, 2017.

\bibitem{xie2012image}
Junyuan Xie, Linli Xu, and Enhong Chen.
\newblock Image denoising and inpainting with deep neural networks.
\newblock {\em Advances in neural information processing systems}, 25, 2012.

\bibitem{liu2018multi}
Pengju Liu, Hongzhi Zhang, Kai Zhang, Liang Lin, and Wangmeng Zuo.
\newblock Multi-level wavelet-cnn for image restoration.
\newblock In {\em Proceedings of the IEEE conference on computer vision and
  pattern recognition workshops}, pages 773--782, 2018.

\bibitem{jia2021ddunet}
Fan Jia, Wing~Hong Wong, and Tieyong Zeng.
\newblock Ddunet: Dense dense u-net with applications in image denoising.
\newblock In {\em Proceedings of the IEEE/CVF International Conference on
  Computer Vision}, pages 354--364, 2021.

\bibitem{hummel1975histogram}
Robert~A Hummel.
\newblock Histogram modification techniques.
\newblock {\em Computer Graphics and Image Processing}, 4(3):209--224, 1975.

\bibitem{gangkofner2007optimizing}
Ute~G Gangkofner, Pushkar~S Pradhan, and Derrold~W Holcomb.
\newblock Optimizing the high-pass filter addition technique for image fusion.
\newblock {\em Photogrammetric Engineering \& Remote Sensing},
  73(9):1107--1118, 2007.

\bibitem{svoboda2016compression}
Pavel Svoboda, Michal Hradis, David Barina, and Pavel Zemcik.
\newblock Compression artifacts removal using convolutional neural networks.
\newblock {\em arXiv preprint arXiv:1605.00366}, 2016.

\bibitem{yu2016deep}
Ke~Yu, Chao Dong, Chen~Change Loy, and Xiaoou Tang.
\newblock Deep convolution networks for compression artifacts reduction.
\newblock {\em arXiv preprint arXiv:1608.02778}, 2016.

\bibitem{galteri2017deep}
Leonardo Galteri, Lorenzo Seidenari, Marco Bertini, and Alberto Del~Bimbo.
\newblock Deep generative adversarial compression artifact removal.
\newblock In {\em Proceedings of the IEEE International Conference on Computer
  Vision}, pages 4826--4835, 2017.

\bibitem{dong2014learning}
Chao Dong, Chen~Change Loy, Kaiming He, and Xiaoou Tang.
\newblock Learning a deep convolutional network for image super-resolution.
\newblock In {\em European conference on computer vision}, pages 184--199.
  Springer, 2014.

\bibitem{alakuijala2017guetzli}
Jyrki Alakuijala, Robert Obryk, Ostap Stoliarchuk, Zoltan Szabadka, Lode
  Vandevenne, and Jan Wassenberg.
\newblock Guetzli: Perceptually guided jpeg encoder.
\newblock {\em arXiv preprint arXiv:1703.04421}, 2017.

\bibitem{wortsman2022model}
Mitchell Wortsman, Gabriel Ilharco, Samir~Ya Gadre, Rebecca Roelofs, Raphael
  Gontijo-Lopes, Ari~S Morcos, Hongseok Namkoong, Ali Farhadi, Yair Carmon,
  Simon Kornblith, et~al.
\newblock Model soups: averaging weights of multiple fine-tuned models improves
  accuracy without increasing inference time.
\newblock In {\em International Conference on Machine Learning}, pages
  23965--23998. PMLR, 2022.

\bibitem{enwiki:1102976887}
{Wikipedia contributors}.
\newblock Permutation --- {Wikipedia}{,} the free encyclopedia.
\newblock
  \url{https://en.wikipedia.org/w/index.php?title=Permutation&oldid=1102976887},
  2022.
\newblock [Online; accessed 9-August-2022].

\bibitem{ronneberger2015u}
Olaf Ronneberger, Philipp Fischer, and Thomas Brox.
\newblock U-net: Convolutional networks for biomedical image segmentation.
\newblock In {\em International Conference on Medical image computing and
  computer-assisted intervention}, pages 234--241. Springer, 2015.

\bibitem{mit-tuebingen-saliency-benchmark}
Matthias K{\"u}mmerer, Zoya Bylinskii, Tilke Judd, Ali Borji, Laurent Itti,
  Fr{\'e}do Durand, Aude Oliva, and Antonio Torralba.
\newblock Mit/tübingen saliency benchmark.
\newblock \url{https://saliency.tuebingen.ai/}.

\bibitem{9746539}
Anne-Flore Perrin, Yejing Xie, Tao Zhang, Yiting Liao, Junlin Li, and
  Patrick~Le Callet.
\newblock Specialised video quality model for enhanced user generated content
  (ugc) with special effects.
\newblock In {\em ICASSP 2022 - 2022 IEEE International Conference on
  Acoustics, Speech and Signal Processing (ICASSP)}, pages 2040--2044, 2022.

\bibitem{ICME}
Challenge on quality assessment of compressed ugc videos.
\newblock Available at \url{http://ugcvqa.com}.

\bibitem{kingma2014adam}
Diederik~P Kingma and Jimmy Ba.
\newblock Adam: A method for stochastic optimization. iclr. 2015.
\newblock {\em arXiv preprint arXiv:1412.6980}, 9, 2015.

\end{thebibliography}
\end{document}